\documentclass{article}
\usepackage{spconf,amsmath,graphicx}

\usepackage{cite}
\usepackage{amsmath,amssymb,amsfonts}
\usepackage{algorithmic}
\usepackage{graphicx}
\usepackage{textcomp}
\usepackage{xcolor}
\usepackage{color}
\usepackage{multirow}
\usepackage{makecell}
\usepackage{marginnote}
\usepackage{subfigure}
\usepackage{hyperref}

\begin{document}

\title{SJTU-TMQA: A quality assessment database for static mesh with texture map}

\name{Bingyang Cui$^{\star}$ \qquad Qi Yang$^{\dagger}$ \qquad Kaifa Yang$^{\star}$ \qquad Yiling Xu$^{\star}$ \qquad Xiaozhong Xu$^{\dagger}$ \qquad Shan Liu$^{\dagger}$}
  
  \address{$^{\star}$ Cooperative Medianet Innovation Center, Shanghai Jiaotong University\\
      $^{\dagger}$Media Lab, Tencent}

\maketitle

\begin{abstract}
In recent years, static meshes with texture maps have become one of the most prevalent digital representations of 3D shapes in various applications, such as animation, gaming, medical imaging, and cultural heritage applications. However, little research has been done on the quality assessment of textured meshes, which hinders the development of quality-oriented applications, such as mesh compression and enhancement. In this paper, we create a large-scale textured mesh quality assessment database, namely SJTU-TMQA, which includes 21 reference meshes and 945 distorted samples. The meshes are rendered into processed video sequences and then conduct subjective experiments to obtain mean opinion scores (MOS). The diversity of content and accuracy of MOS has been shown to validate its heterogeneity and reliability. The impact of various types of distortion on human perception is demonstrated. 13 state-of-the-art objective metrics are evaluated on SJTU-TMQA. The results report the highest correlation of around 0.6, indicating the need for more effective objective metrics. The SJTU-TMQA is available at \href{https://ccccby.github.io}{https://ccccby.github.io}
\end{abstract}


\section{Introduction}
With the technological advancements of computer graphics and the development of rendering technologies, 3D static meshes with texture maps are constantly applied in many areas due to their effectiveness in representing 3D objects or scenes. A typical 3D textured mesh contains a number of faces with 3D points as vertices, each face is textured with a texture map indicated by texture coordinates. For brevity, we use textured mesh to indicate static mesh with texture map. The quality of textured mesh is important for human perception-oriented applications, such as immersive gaming, animation, and digital museums. However, 3D textured meshes have a large volume of data. They require effective compression and transmission algorithms before practical utilizations, in which different types of distortion might be introduced and degrade subjective perceived quality. To optimize textured mesh processing algorithms with respect to quality of experience, mesh quality assessment (MQA) has become a hotspot in recent study \cite{tsmd,Nehm2022dataset, Nehmvertexdataset}. 

MQA includes two aspects: subjective and objective quality assessment. Subjective quality assessment is the most reliable method, which needs to invite subjects to evaluate the perceptual quality of distorted meshes in strictly controlled testing environments. Objective quality assessment aims to study objective metrics that have high correlations with human perceptual quality, replacing subjective experiments in practical and real-time applications to reduce the cost of time, human resources, and money. Therefore, to design effective objective quality metrics and facilitate the application of textured meshes, subjective MQA needs to be fully studied, and a database containing diverse mesh contents, rich distortion types, and reliable mean opinion scores (MOS) is expected. 

Over the past years, some researchers have conducted studies on subjective MQA and established several databases. For example, \cite{guillaume2006perceptually, guillaume2009roughness} focus on colorless meshes and mainly consider single distortion types, such as noise addition and lossy compression. \cite{Nehmvertexdataset} studies meshes with vertex color and releases a database with 480 distorted meshes under compression and simplification distortion. \cite{tsmd, Nehm2022dataset} investigate textured meshes and propose superimposed distortion types, including mesh simplification/decimation, texture map downsampling, and coordinate quantization. 

However, the aforementioned public databases have weaknesses, limiting their utilization in current studies. First, \cite{guillaume2006perceptually, guillaume2009roughness, Nehmvertexdataset} are for colorless or vertex-color meshes, while meshes with texture map are the star of emerging immersive multimedia applications. Second, they are limited by the small-scale \cite{guillaume2006perceptually, guillaume2009roughness} or the restricted range of distortion types \cite{Nehmvertexdataset, tsmd, Nehm2022dataset}, making them insufficient for a comprehensive MQA study. 

To mitigate the above problems, we create a large-scale textured mesh database containing rich contents and multiple types of distortion in this paper, called SJTU-TMQA. 21 reference meshes are selected from different categories, including human figures, inanimate objects, animals, and plants. Eight types of distortion: six single distortion types and two superimposed distortion types are injected into each reference mesh at different distortion levels, leading to 945 distorted meshes. The distorted meshes are rendered into processed video sequences (PVS) with a predefined camera path, and 73 viewers aged 18 to 30 are collected to perform subjective experiments with a lab environment. The diversity of source content, the accuracy of the MOS, and the influence of different types of distortion are demonstrated. 13 state-of-the-art (SOTA) objective metrics are tested on SJTU-TMQA. The best results report correlations of around 0.60, indicating that the proposed SJTU-TMQA is a challenging database and serves as a catalyst for a more effective objective metric study.

\section{Database Construction}\label{2}
In this section, we detail the construction of SJTU-TMQA, including source mesh selection, distortion generation, PVS generation, training and rating session, and outlier removal. 

\subsection{Source mesh selection and preprocessing}
\begin{figure}[htbp]
\setlength{\abovecaptionskip}{0.cm}
\setlength{\belowcaptionskip}{-0.cm}
\centerline{\includegraphics[width=0.4\textwidth]{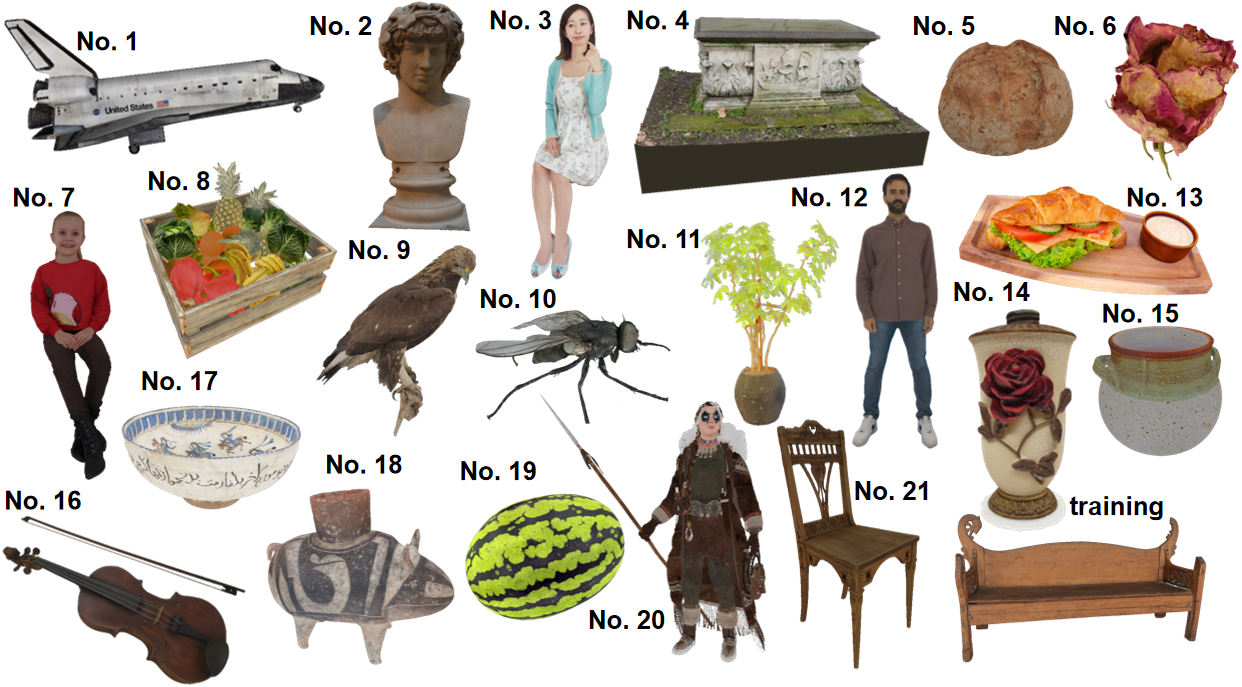}}
\caption{The 3D graphic source model of our database}
\label{mainviewpoint_image}
\vspace{-2.0em}
\end{figure}

To better study the perceived subjective quality of textured meshes, 21 high quality source meshes are carefully selected from SketchFab\footnote{https://sketchfab.com/features/free-3d-models}. These meshes encompass a diverse array of categories, including human figures, inanimate objects, animals, and plants. Fig. \ref{mainviewpoint_image} illustrates the snapshots of the source content. PymeshLab\footnote{https://github.com/cnr-isti-vclab/PyMeshLab} library is used to remove redundant and invalid information (e.g., unreferenced vertices and null faces) from the reference mesh as proposed in \cite{yang2023geodesicpsim}. 

\subsection{Distortion generation}
To simulate various types of distortion resulting from acquisition noise, resampling, compression, and other factors,  8 different distortion types are introduced and detailed as follows:

    $\bullet$\textbf{Downsampling (DS)}: DS is applied to the texture map of the textured mesh. The ``Image.LANCZOS" low-pass filter offered by PIL library\footnote{https://github.com/python-pillow/Pillow} is used to resize the texture map to 45\%, 35\%, 25\%, 15\%, and 5\% of the original resolution.
    
    $\bullet$\textbf{Gaussion noise (GN)}: GN is applied to the vertex coordinates of the textured mesh.  All vertices of reference meshes are enhanced with a random Gaussian distributed geometry shift which magnitude are 0.5\%, 1.0\%, 1.5\%, 2.0\%, and 2.5\% of the minimum dimension of the bounding box.
    
    $\bullet$\textbf{Texture map compression (TMC)}: TMC is applied to the texture map of the textured mesh. We use the ``imwrite(`jpg', `Quality')" compression function offered by Matlab software, which is based on the libjpeg library\footnote{https://jpeg.org/jpeg/software.html}, with the following quality parameters: 24, 20, 16, 12, 8, and 4. 
    
    $\bullet$\textbf{Quantization Position (QP)}: QP is applied to the vertex coordinates of the textured mesh. Draco\footnote{https://github.com/google/draco} is used to perform uniform quantization with bits set to 7, 8, 9, 10, and 11.

    $\bullet$\textbf{Simplification without texture (SOT)}: SOT is applied to the faces of the mesh sample, in which the number of vertices is reduced and consequently leads to larger face sizes.  Iterative edge collapse and a quadric error metric (QEM) \cite{cignoni2008meshlab} are used to perform simplification and reduce the number of faces by 10\%, 25\%, 40\%, and 55\% compared to source meshes.

    $\bullet$\textbf{Simplification with texture (SWT)}: SWT is also applied to the faces of the mesh sample, but the texture information is injected to guide the QEM simplification results. We uniformly reduce the number of faces by 20\%, 35\%, 50\%, 65\%, and 80\% compared to source meshes.

    $\bullet$\textbf{Mixed Quantization (MQ)}: MQ is a superimposed distortion that applied QP and QT (texture coordinate quantization in Draco) at the same time. We carefully set the appropriate parameters, i.e. (QP / bits, QT / bits), to (12, 12), (11, 12), (10, 12), (9, 11), (8, 10), (7, 8).

    $\bullet$\textbf{Geometry and Texture map Compression (GTC)}: GTC is a superimposed distortion which is a combination of MQ and TMC distortion. We selected three distortion levels from MQ ((11, 12), (9, 11), and (7, 8)) and TMC distortion (20, 12, and 4), respectively, leading to the generation of 3x3=9 distorted meshes with pair matching.

 
 In all, we obtain $ \rm 21\ x\ (5+5+6+5+4+5+6+9)\ = 945$ distorted meshes.

\subsection{PVS generation}\label{4}

To perform subjective experiments,  each distorted mesh is rendered to PVS with 1920x1080 resolution and 30 fps, using a pre-defined camera paths: the camera rotates around the $z$ axis with a rotation step of $0.75{^{\circ}}$ degrees per frame, and the rotation radius is equal to the mesh maximum bounding box.  A complete rotation ($360{^{\circ}}$) around the mesh results in 495 frame images captured by OpenGL. Then, we group the images into PVSs using FFMPEG with libx265 , and the constant rate factor is set to 10 to ensure visually lossless encoding \cite{yang2023exploring}. Each PVS has a duration of 16 seconds.

\subsection{Training and rating session}
To ensure the reliability of the collected subjective scores, we use ``bench'' shown in Fig. \ref{mainviewpoint_image} to generate a training session with the same method as \cite{tsmd}. In the rating session, a double stimulus impairment scale method is used and an 11-level impairment scale proposed by ITU-T P. 910 \cite{p910} is used as the voting method. The subjective experiment is conducted on a 27-inch AOC Q2790PQ monitor with resolution 2560×1440 in an indoor lab environment under normal lighting conditions. The display resolution is adjusted to 1920×1080 to ensure the consistency with the PVSs. To avoid visual fatigue caused by an overly long experiment time, we randomly divide the 945 PVS into 21 subgroups. 

\subsection{Outlier removal}

Two consecutive steps are adopted to remove outliers from the raw subjective scores. First, each rating session additionally contains an extremely low-quality PVS and a duplicated PVS, known as "trapping samples". After collecting subjective scores, we first remove outliers according to the trapping results. Second,  ITU-R BT.500 \cite{bt500} is used to detect and remove outliers again. Finally, three outliers are identified and removed from the poor subjective score. 

\section{database analysis}\label{5}
In this section, the diversity of content in SJTU-TMQA is first proved, then subjective experiment results are analyzed to demonstrate the reliability of MOS.

\subsection{Diversity of SJTU-TMQA content}
\begin{figure}[h]
\setlength{\abovecaptionskip}{0.cm}
\setlength{\belowcaptionskip}{-0.cm}
	\centering
 \subfigure[]{\includegraphics[width=0.48\linewidth]{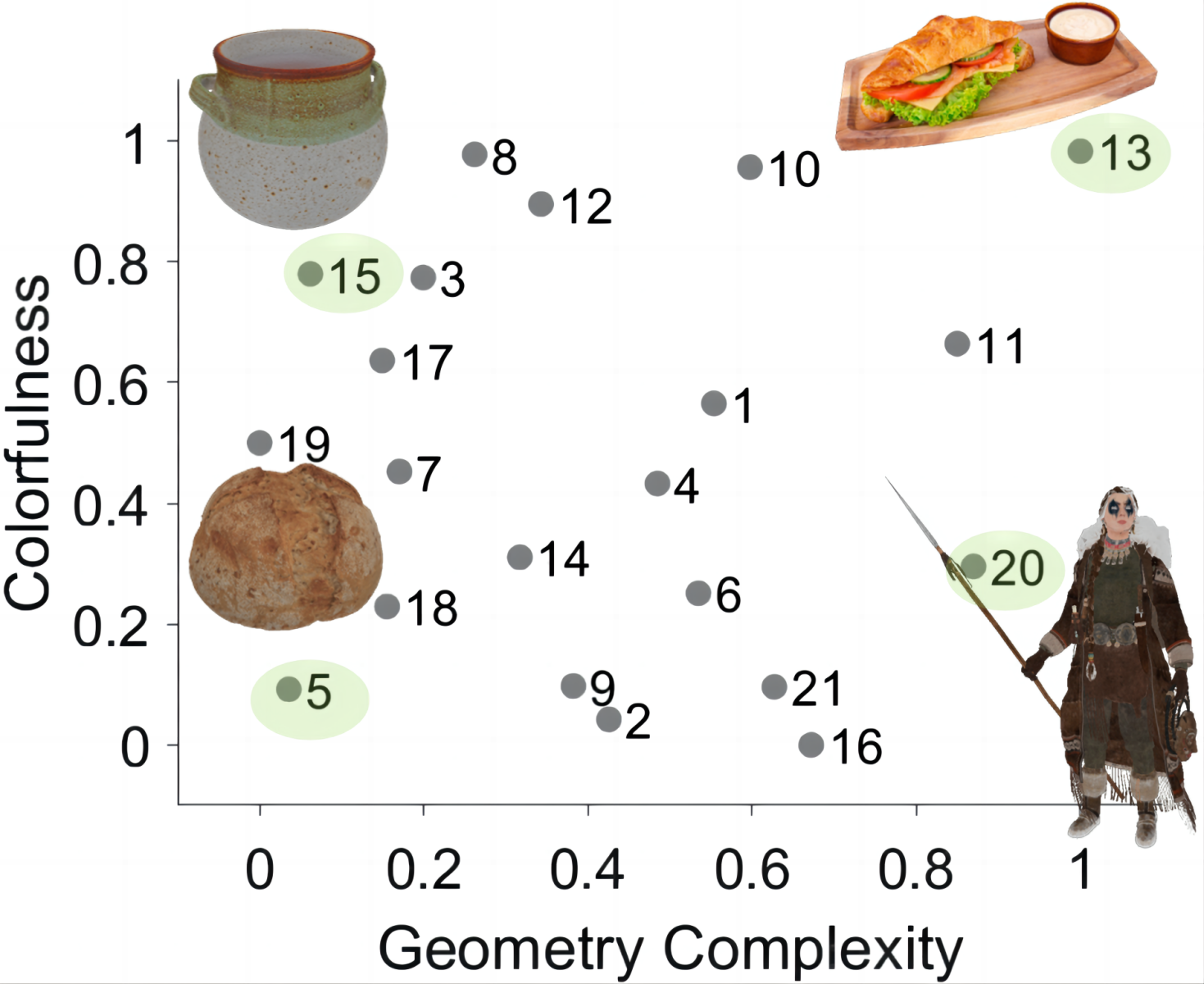}\label{fig:complexity}}
    \subfigure[]{\includegraphics[width=0.48\linewidth]{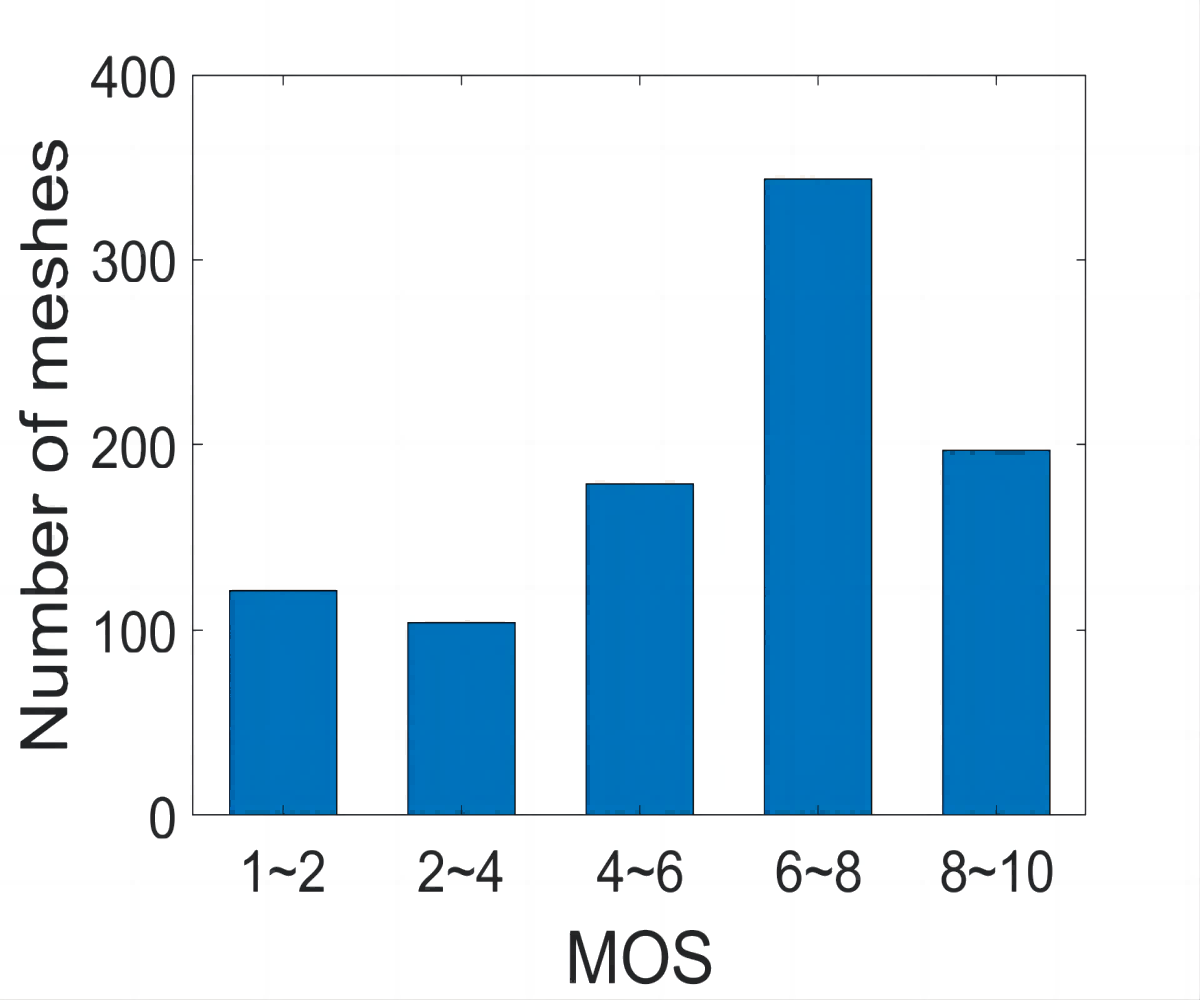}\label{fig:mos_dis}}
	\caption{(a): Geometry vs. Color complexity; (b): MOS distribution of SJTU-TMQA}
	\label{fig:bivsmos}
  \vspace{-2.0em}
\end{figure}

Geometry and color complexities are proposed to validate the diversity of content, which quantified by spatial perceptual information (SI) \cite{p910} and the color metric (CM) \cite{colorCF-hasler2003measuring}, respectively. We use the depth and color image obtained by projection with six views of its bounding box \cite{yang2020predicting} to calculate the SI and CM of the reference mesh. The maximum SI and CM values are selected to illustrate the scatter plot of geometry complexity vs. color complexity in Fig. \ref{fig:complexity}. The relatively uniform distribution of the scatter points indicates the diversity of the SJTU-TMQA content. 

\subsection{Analysis of MOS}

Fig. \ref{fig:mos_dis} reports the MOS distribution of SJTU-TMQA. For each score segment, SJTU-TMQA has at least 100 distorted meshes, indicating that SJTU-TMQA covers a wide range of quality scores. 
\begin{figure}[pt]
\setlength{\abovecaptionskip}{-1.cm}
\setlength{\belowcaptionskip}{-1.cm}
	\centering
	\subfigure{		\includegraphics[width=0.3\linewidth]{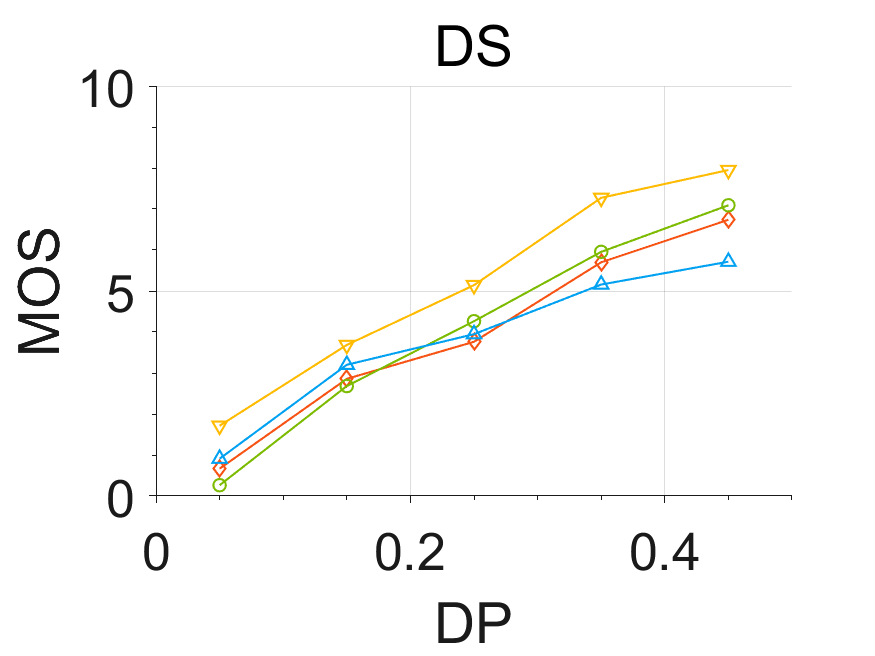}			}%
	\subfigure{		\includegraphics[width=0.3\linewidth]{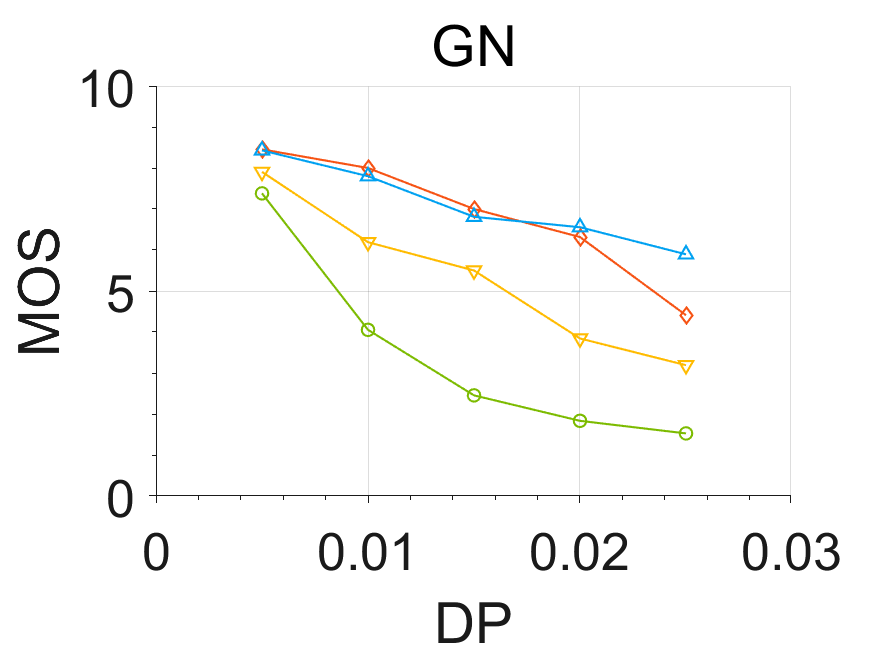}			}
	\subfigure{		\includegraphics[width=0.3\linewidth]{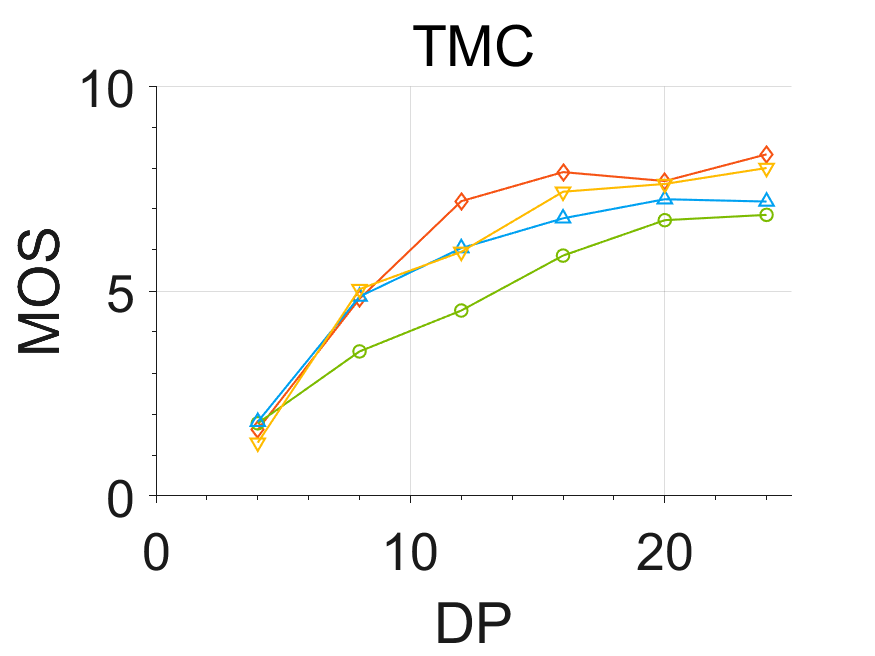}			}\\
	\subfigure{		\includegraphics[width=0.3\linewidth]{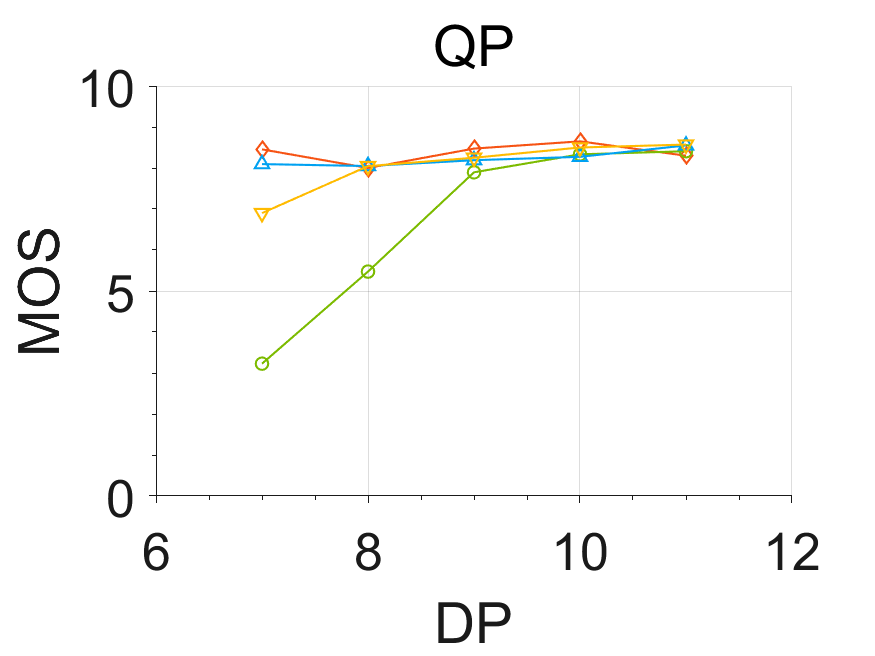}			} 
	\subfigure{		\includegraphics[width=0.3\linewidth]{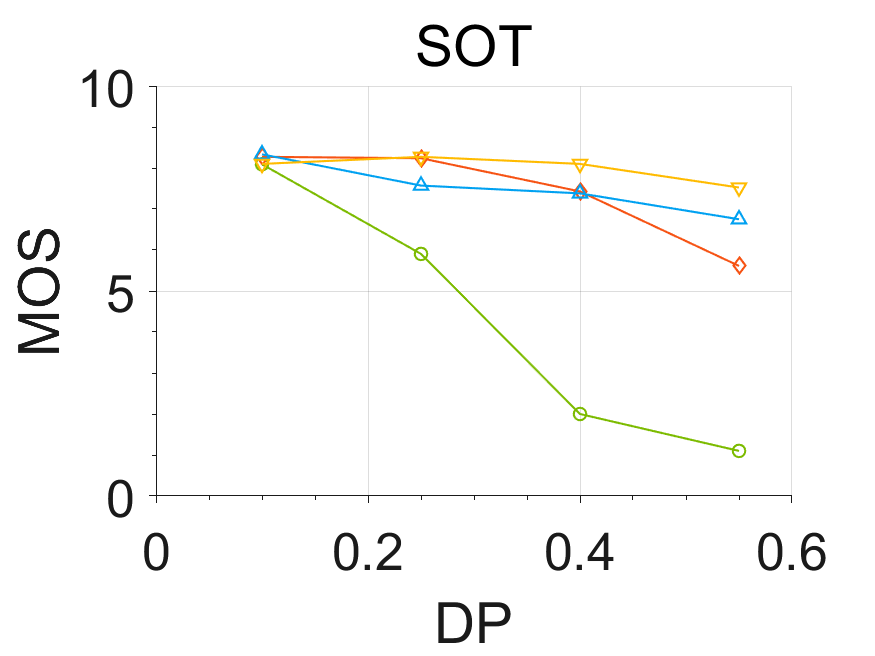}			}%
	\subfigure{		\includegraphics[width=0.3\linewidth]{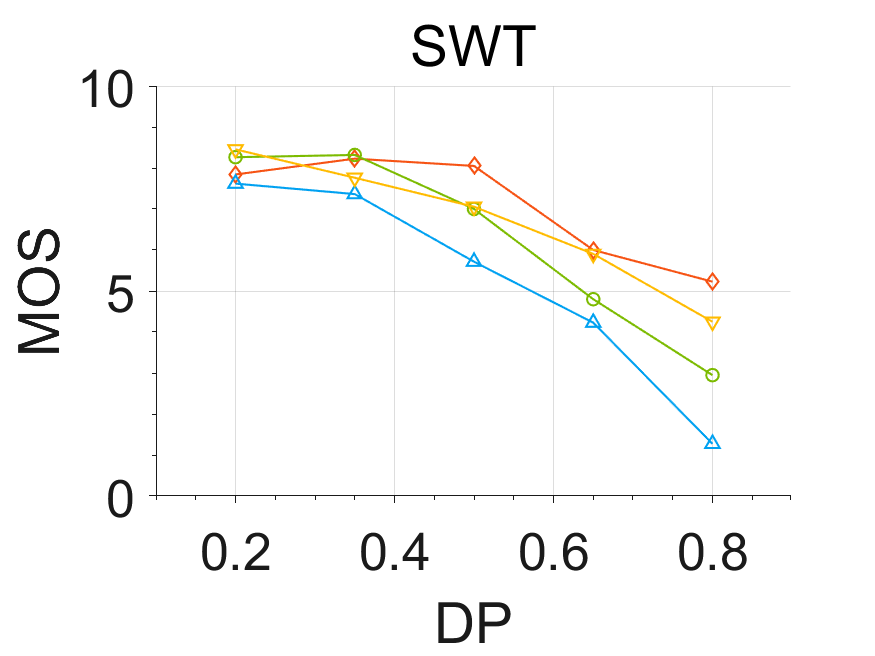}			}\\
     \subfigure{		\includegraphics[width=0.6\linewidth]{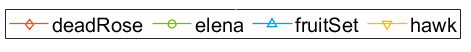}			}
	\caption{MOS vs. Distortion Parameters (DP).}
	\label{fig:mosvsdl}
 \vspace{-2em}
\end{figure}

To prove the accuracy of MOS and analyze the impact of different distortion on subjective perception, MOS vs. distortion parameter plots of four meshes which belong to different types of content (i.e., deadRose, elena, fruitSet, and hawk), are shown in Fig. \ref{fig:mosvsdl}. Except for QP, most of the curves of DS, GN, TMC, SOT, and SWT showcase perfect monotonicity, which proves the accuracy of the MOS. For QP, except for ``elena'', the other three meshes present limited MOS variations. We think the reasons are: first, the influence of QP can be masked by mesh texture; and second, ``elena'' belongs to the human figure and human observers are particularly sensitive to facial features that are known as salient areas \cite{yang2023exploring}. Minor distortion in these areas can easily be detected and reflected via MOS variation.

\section{Objective metrics testing}
Four types of objective metrics are tested based on SJTU-TMQA: image-based, point-based, video-based, and model-based metrics. Image-based metrics, proposed by\cite{wg7},  use 16 projected images of meshes to quantify quality. Two image-based quality metrics (Geo\_PSNR and RGB\_PSNR) are tested. Point-based metrics first use sampling to convert mesh into point clouds, and then measure quality using point cloud objective metrics. Four point-based metrics (D1 \cite{d1}, D2 \cite{d2}, YUV\_PSNR, and PCQM\_PSNR \cite{pcqm_psnr}) are tested. Grid sampling with a grid resolution of 1024 is used to sample meshes into point clouds as proposed in \cite{wg7}. Video-based metrics use the PVSs viewed in the subjective experiment as input, then image/video quality metrics are applied to predict mesh quality. Three video-based metrics (PSNR, SSIM \cite{ssim}, VMAF \cite{vmaf}) are calculated. Model-based metrics directly use the raw data from the mesh to assess quality. Four model-based metrics (Hausedorff distance (HD) \cite{hausedorff}, GL2 \cite{gl1/2}, MSDM2 \cite{msdm2}, and TPDM \cite{tpdm}) are tested. 

\subsection{Performance of metrics}

To ensure consistency between the objective and MOS of the various objective metrics, a five-parameter logistic fitting function proposed by the video quality experts group \cite{fittingfunction} is used to map the dynamic range of the scores from the objective metric to a common scale. Two indicators commonly used in quality assessment society are offered to quantify the efficiency of various metrics: Pearson linear correlation coefficient (PLCC) for prediction accuracy, and Spearman rank-order correlation coefficient (SRCC) for prediction monotonicity. 

\subsection{Correlation of metric}
\begin{table}[]
\caption{Metric performance on SJTU-TMQA}
\label{tb:metrics_performance}
\small
  \begin{scriptsize}
 \renewcommand{\arraystretch}{0.5}
	\setlength{\tabcolsep}{0.5mm}{
\begin{tabular}{|cc|cc|cccccccc|}
\hline
\multicolumn{2}{|c|}{} & \multicolumn{2}{c|}{} & \multicolumn{8}{c|}{} \\
\multicolumn{2}{|c|}{\multirow{-2}{*}{Index}} & \multicolumn{2}{c|}{\multirow{-2}{*}{All}} & \multicolumn{8}{c|}{\multirow{-2}{*}{Distortion}} \\ \hline
\multicolumn{1}{|c|}{} &  & \multicolumn{1}{c|}{} &  & \multicolumn{1}{c|}{} & \multicolumn{1}{c|}{} & \multicolumn{1}{c|}{} & \multicolumn{1}{c|}{} & \multicolumn{1}{c|}{} & \multicolumn{1}{c|}{} & \multicolumn{1}{c|}{} &  \\
\multicolumn{1}{|c|}{\multirow{-2}{*}{Type}} & \multirow{-2}{*}{Metric} & \multicolumn{1}{c|}{\multirow{-2}{*}{PLCC}} & \multirow{-2}{*}{SRCC} & \multicolumn{1}{c|}{\multirow{-2}{*}{DS}} & \multicolumn{1}{c|}{\multirow{-2}{*}{GN}} & \multicolumn{1}{c|}{\multirow{-2}{*}{TMC}} & \multicolumn{1}{c|}{\multirow{-2}{*}{QP}} & \multicolumn{1}{c|}{\multirow{-2}{*}{SWT}} & \multicolumn{1}{c|}{\multirow{-2}{*}{SOT}} & \multicolumn{1}{c|}{\multirow{-2}{*}{MQ}} & \multirow{-2}{*}{GTC} \\ \hline
\multicolumn{1}{|c|}{} &  & \multicolumn{1}{c|}{} &  & \multicolumn{1}{c|}{} & \multicolumn{1}{c|}{} & \multicolumn{1}{c|}{} & \multicolumn{1}{c|}{} & \multicolumn{1}{c|}{} & \multicolumn{1}{c|}{} & \multicolumn{1}{c|}{} &  \\
\multicolumn{1}{|c|}{} & \multirow{-2}{*}{Geo\_PSNR} & \multicolumn{1}{c|}{\multirow{-2}{*}{0.16}} & \multirow{-2}{*}{0.09} & \multicolumn{1}{c|}{\multirow{-2}{*}{-}} & \multicolumn{1}{c|}{\multirow{-2}{*}{0.41}} & \multicolumn{1}{c|}{\multirow{-2}{*}{-}} & \multicolumn{1}{c|}{\multirow{-2}{*}{0.62}} & \multicolumn{1}{c|}{\multirow{-2}{*}{0.46}} & \multicolumn{1}{c|}{\multirow{-2}{*}{0.30}} & \multicolumn{1}{c|}{\multirow{-2}{*}{0.69}} & \multirow{-2}{*}{0.20} \\ \cline{2-12} 
\multicolumn{1}{|c|}{} &  & \multicolumn{1}{c|}{} &  & \multicolumn{1}{c|}{} & \multicolumn{1}{c|}{} & \multicolumn{1}{c|}{{\color[HTML]{333333} }} & \multicolumn{1}{c|}{} & \multicolumn{1}{c|}{} & \multicolumn{1}{c|}{} & \multicolumn{1}{c|}{} &  \\
\multicolumn{1}{|c|}{\multirow{-4}{*}{A}} & \multirow{-2}{*}{RGB\_PSNR} & \multicolumn{1}{c|}{\multirow{-2}{*}{0.55}} & \multirow{-2}{*}{0.58} & \multicolumn{1}{c|}{\multirow{-2}{*}{0.74}} & \multicolumn{1}{c|}{\multirow{-2}{*}{0.35}} & \multicolumn{1}{c|}{\multirow{-2}{*}{{\color[HTML]{333333} 0.62}}} & \multicolumn{1}{c|}{\multirow{-2}{*}{0.64}} & \multicolumn{1}{c|}{\multirow{-2}{*}{0.51}} & \multicolumn{1}{c|}{\multirow{-2}{*}{0.63}} & \multicolumn{1}{c|}{\multirow{-2}{*}{0.67}} & \multirow{-2}{*}{0.57} \\ \hline
\multicolumn{1}{|c|}{} &  & \multicolumn{1}{c|}{{\color[HTML]{5B9BD5} }} &  & \multicolumn{1}{c|}{} & \multicolumn{1}{c|}{} & \multicolumn{1}{c|}{} & \multicolumn{1}{c|}{{\color[HTML]{FF0000} }} & \multicolumn{1}{c|}{{\color[HTML]{FF0000} }} & \multicolumn{1}{c|}{} & \multicolumn{1}{c|}{{\color[HTML]{FF0000} }} &  \\
\multicolumn{1}{|c|}{} & \multirow{-2}{*}{D1} & \multicolumn{1}{c|}{\multirow{-2}{*}{{\color[HTML]{5B9BD5} \textbf{0.05}}}} & \multirow{-2}{*}{0.13} & \multicolumn{1}{c|}{\multirow{-2}{*}{-}} & \multicolumn{1}{c|}{\multirow{-2}{*}{0.46}} & \multicolumn{1}{c|}{\multirow{-2}{*}{-}} & \multicolumn{1}{c|}{\multirow{-2}{*}{{\color[HTML]{FF0000} \textbf{0.76}}}} & \multicolumn{1}{c|}{\multirow{-2}{*}{0.66}} & \multicolumn{1}{c|}{\multirow{-2}{*}{0.45}} & \multicolumn{1}{c|}{\multirow{-2}{*}{{\color[HTML]{FF0000} \textbf{0.79}}}} & \multirow{-2}{*}{0.31} \\ \cline{2-12} 
\multicolumn{1}{|c|}{} &  & \multicolumn{1}{c|}{} &  & \multicolumn{1}{c|}{} & \multicolumn{1}{c|}{} & \multicolumn{1}{c|}{} & \multicolumn{1}{c|}{{\color[HTML]{FF0000} }} & \multicolumn{1}{c|}{} & \multicolumn{1}{c|}{} & \multicolumn{1}{c|}{{\color[HTML]{FF0000} }} &  \\
\multicolumn{1}{|c|}{} & \multirow{-2}{*}{D2} & \multicolumn{1}{c|}{\multirow{-2}{*}{0.09}} & \multirow{-2}{*}{0.13} & \multicolumn{1}{c|}{\multirow{-2}{*}{-}} & \multicolumn{1}{c|}{\multirow{-2}{*}{0.43}} & \multicolumn{1}{c|}{\multirow{-2}{*}{-}} & \multicolumn{1}{c|}{\multirow{-2}{*}{0.75}} & \multicolumn{1}{c|}{\multirow{-2}{*}{0.64}} & \multicolumn{1}{c|}{\multirow{-2}{*}{0.40}} & \multicolumn{1}{c|}{\multirow{-2}{*}{0.78}} & \multirow{-2}{*}{0.30} \\ \cline{2-12} 
\multicolumn{1}{|c|}{} &  & \multicolumn{1}{c|}{{\color[HTML]{FF0000} }} & {\color[HTML]{FF0000} } & \multicolumn{1}{c|}{} & \multicolumn{1}{c|}{} & \multicolumn{1}{c|}{} & \multicolumn{1}{c|}{} & \multicolumn{1}{c|}{} & \multicolumn{1}{c|}{} & \multicolumn{1}{c|}{} & {\color[HTML]{FF0000} } \\
\multicolumn{1}{|c|}{} & \multirow{-2}{*}{YUV\_PSNR} & \multicolumn{1}{c|}{\multirow{-2}{*}{{\color[HTML]{FF0000} \textbf{0.59}}}} & \multirow{-2}{*}{{\color[HTML]{FF0000} \textbf{0.65}}} & \multicolumn{1}{c|}{\multirow{-2}{*}{0.78}} & \multicolumn{1}{c|}{\multirow{-2}{*}{0.37}} & \multicolumn{1}{c|}{\multirow{-2}{*}{0.64}} & \multicolumn{1}{c|}{\multirow{-2}{*}{0.66}} & \multicolumn{1}{c|}{\multirow{-2}{*}{0.49}} & \multicolumn{1}{c|}{\multirow{-2}{*}{0.68}} & \multicolumn{1}{c|}{\multirow{-2}{*}{0.67}} & \multirow{-2}{*}{{\color[HTML]{FF0000} \textbf{0.59}}} \\ \cline{2-12} 
\multicolumn{1}{|c|}{} &  & \multicolumn{1}{c|}{} &  & \multicolumn{1}{c|}{} & \multicolumn{1}{c|}{} & \multicolumn{1}{c|}{} & \multicolumn{1}{c|}{} & \multicolumn{1}{c|}{} & \multicolumn{1}{c|}{{\color[HTML]{FF0000} }} & \multicolumn{1}{c|}{} &  \\
\multicolumn{1}{|c|}{\multirow{-8}{*}{B}} & \multirow{-2}{*}{PCQM\_PSNR} & \multicolumn{1}{c|}{\multirow{-2}{*}{0.48}} & \multirow{-2}{*}{0.55} & \multicolumn{1}{c|}{\multirow{-2}{*}{0.78}} & \multicolumn{1}{c|}{\multirow{-2}{*}{0.29}} & \multicolumn{1}{c|}{\multirow{-2}{*}{0.61}} & \multicolumn{1}{c|}{\multirow{-2}{*}{0.68}} & \multicolumn{1}{c|}{\multirow{-2}{*}{0.52}} & \multicolumn{1}{c|}{\multirow{-2}{*}{{\color[HTML]{FF0000} \textbf{0.70}}}} & \multicolumn{1}{c|}{\multirow{-2}{*}{0.69}} & \multirow{-2}{*}{0.43} \\ \hline
\multicolumn{1}{|c|}{} &  & \multicolumn{1}{c|}{} &  & \multicolumn{1}{c|}{} & \multicolumn{1}{c|}{} & \multicolumn{1}{c|}{{\color[HTML]{5B9BD5} }} & \multicolumn{1}{c|}{} & \multicolumn{1}{c|}{} & \multicolumn{1}{c|}{} & \multicolumn{1}{c|}{} &  \\
\multicolumn{1}{|c|}{} & \multirow{-2}{*}{PSNR} & \multicolumn{1}{c|}{\multirow{-2}{*}{0.40}} & \multirow{-2}{*}{0.44} & \multicolumn{1}{c|}{\multirow{-2}{*}{0.73}} & \multicolumn{1}{c|}{\multirow{-2}{*}{0.38}} & \multicolumn{1}{c|}{\multirow{-2}{*}{{\color[HTML]{5B9BD5} \textbf{0.46}}}} & \multicolumn{1}{c|}{\multirow{-2}{*}{0.66}} & \multicolumn{1}{c|}{\multirow{-2}{*}{0.51}} & \multicolumn{1}{c|}{\multirow{-2}{*}{0.48}} & \multicolumn{1}{c|}{\multirow{-2}{*}{0.68}} & \multirow{-2}{*}{0.05} \\ \cline{2-12} 
\multicolumn{1}{|c|}{} &  & \multicolumn{1}{c|}{} &  & \multicolumn{1}{c|}{{\color[HTML]{5B9BD5} }} & \multicolumn{1}{c|}{} & \multicolumn{1}{c|}{} & \multicolumn{1}{c|}{} & \multicolumn{1}{c|}{} & \multicolumn{1}{c|}{} & \multicolumn{1}{c|}{} & {\color[HTML]{5B9BD5} } \\
\multicolumn{1}{|c|}{} & \multirow{-2}{*}{SSIM} & \multicolumn{1}{c|}{\multirow{-2}{*}{0.33}} & \multirow{-2}{*}{0.48} & \multicolumn{1}{c|}{\multirow{-2}{*}{{\color[HTML]{5B9BD5} \textbf{0.69}}}} & \multicolumn{1}{c|}{\multirow{-2}{*}{0.03}} & \multicolumn{1}{c|}{\multirow{-2}{*}{0.50}} & \multicolumn{1}{c|}{\multirow{-2}{*}{0.61}} & \multicolumn{1}{c|}{\multirow{-2}{*}{0.40}} & \multicolumn{1}{c|}{\multirow{-2}{*}{0.47}} & \multicolumn{1}{c|}{\multirow{-2}{*}{0.67}} & \multirow{-2}{*}{{\color[HTML]{5B9BD5} \textbf{0.01}}} \\ \cline{2-12} 
\multicolumn{1}{|c|}{} &  & \multicolumn{1}{c|}{} &  & \multicolumn{1}{c|}{{\color[HTML]{FF0000} }} & \multicolumn{1}{c|}{{\color[HTML]{333333} }} & \multicolumn{1}{c|}{} & \multicolumn{1}{c|}{} & \multicolumn{1}{c|}{} & \multicolumn{1}{c|}{} & \multicolumn{1}{c|}{} &  \\
\multicolumn{1}{|c|}{\multirow{-6}{*}{C}}& \multirow{-2}{*}{VMAF} & \multicolumn{1}{c|}{\multirow{-2}{*}{0.48}} & \multirow{-2}{*}{0.53} & \multicolumn{1}{c|}{\multirow{-2}{*}{{\color[HTML]{FF0000} \textbf{0.85}}}} & \multicolumn{1}{c|}{\multirow{-2}{*}{{\color[HTML]{333333} 0.48}}} & \multicolumn{1}{c|}{\multirow{-2}{*}{{\color[HTML]{FF0000} \textbf{0.65}}}} & \multicolumn{1}{c|}{\multirow{-2}{*}{0.73}} & \multicolumn{1}{c|}{\multirow{-2}{*}{0.60}} & \multicolumn{1}{c|}{\multirow{-2}{*}{0.48}} & \multicolumn{1}{c|}{\multirow{-2}{*}{0.76}} & \multirow{-2}{*}{0.24} \\ \hline
\multicolumn{1}{|c|}{} &  & \multicolumn{1}{c|}{} &  & \multicolumn{1}{c|}{} & \multicolumn{1}{c|}{} & \multicolumn{1}{c|}{} & \multicolumn{1}{c|}{} & \multicolumn{1}{c|}{} & \multicolumn{1}{c|}{} & \multicolumn{1}{c|}{} &  \\
\multicolumn{1}{|c|}{} & \multirow{-2}{*}{HD} & \multicolumn{1}{c|}{\multirow{-2}{*}{0.14}} & \multirow{-2}{*}{0.06} & \multicolumn{1}{c|}{\multirow{-2}{*}{-}} & \multicolumn{1}{c|}{\multirow{-2}{*}{0.13}} & \multicolumn{1}{c|}{\multirow{-2}{*}{-}} & \multicolumn{1}{c|}{\multirow{-2}{*}{0.12}} & \multicolumn{1}{c|}{\multirow{-2}{*}{0.18}} & \multicolumn{1}{c|}{\multirow{-2}{*}{0.14}} & \multicolumn{1}{c|}{\multirow{-2}{*}{0.24}} & \multirow{-2}{*}{0.09} \\ \cline{2-12} 
\multicolumn{1}{|c|}{} &  & \multicolumn{1}{c|}{} &  & \multicolumn{1}{c|}{} & \multicolumn{1}{c|}{{\color[HTML]{5B9BD5} }} & \multicolumn{1}{c|}{} & \multicolumn{1}{c|}{{\color[HTML]{5B9BD5} }} & \multicolumn{1}{c|}{{\color[HTML]{5B9BD5} }} & \multicolumn{1}{c|}{} & \multicolumn{1}{c|}{} &  \\
\multicolumn{1}{|c|}{} & \multirow{-2}{*}{GL2} & \multicolumn{1}{c|}{\multirow{-2}{*}{0.06}} & \multirow{-2}{*}{0.08} & \multicolumn{1}{c|}{\multirow{-2}{*}{-}} & \multicolumn{1}{c|}{\multirow{-2}{*}{{\color[HTML]{5B9BD5} \textbf{0.01}}}} & \multicolumn{1}{c|}{\multirow{-2}{*}{-}} & \multicolumn{1}{c|}{\multirow{-2}{*}{{\color[HTML]{5B9BD5} \textbf{0.11}}}} & \multicolumn{1}{c|}{\multirow{-2}{*}{{\color[HTML]{5B9BD5} \textbf{0.14}}}} & \multicolumn{1}{c|}{\multirow{-2}{*}{{\color[HTML]{5B9BD5} \textbf{0.03}}}} & \multicolumn{1}{c|}{\multirow{-2}{*}{{\color[HTML]{5B9BD5} \textbf{0.23}}}} & \multirow{-2}{*}{0.08} \\ \cline{2-12} 
\multicolumn{1}{|c|}{} &  & \multicolumn{1}{c|}{} & {\color[HTML]{5B9BD5} } & \multicolumn{1}{c|}{} & \multicolumn{1}{c|}{} & \multicolumn{1}{c|}{} & \multicolumn{1}{c|}{} & \multicolumn{1}{c|}{} & \multicolumn{1}{c|}{} & \multicolumn{1}{c|}{} &  \\
\multicolumn{1}{|c|}{} & \multirow{-2}{*}{MSDM2} & \multicolumn{1}{c|}{\multirow{-2}{*}{0.12}} & \multirow{-2}{*}{{\color[HTML]{5B9BD5} \textbf{0.05}}} & \multicolumn{1}{c|}{\multirow{-2}{*}{-}} & \multicolumn{1}{c|}{\multirow{-2}{*}{0.36}} & \multicolumn{1}{c|}{\multirow{-2}{*}{-}} & \multicolumn{1}{c|}{\multirow{-2}{*}{0.49}} & \multicolumn{1}{c|}{\multirow{-2}{*}{0.64}} & \multicolumn{1}{c|}{\multirow{-2}{*}{0.05}} & \multicolumn{1}{c|}{\multirow{-2}{*}{0.51}} & \multirow{-2}{*}{0.17} \\ \cline{2-12} 
\multicolumn{1}{|c|}{} &  & \multicolumn{1}{c|}{} &  & \multicolumn{1}{c|}{} & \multicolumn{1}{c|}{{\color[HTML]{FF0000} }} & \multicolumn{1}{c|}{} & \multicolumn{1}{c|}{} & \multicolumn{1}{c|}{{\color[HTML]{FF0000} }} & \multicolumn{1}{c|}{} & \multicolumn{1}{c|}{} &  \\
\multicolumn{1}{|c|}{\multirow{-8}{*}{D}} & \multirow{-2}{*}{TPDM} & \multicolumn{1}{c|}{\multirow{-2}{*}{0.15}} & \multirow{-2}{*}{0.10} & \multicolumn{1}{c|}{\multirow{-2}{*}{-}} & \multicolumn{1}{c|}{\multirow{-2}{*}{{\color[HTML]{FF0000} \textbf{0.77}}}} & \multicolumn{1}{c|}{\multirow{-2}{*}{-}} & \multicolumn{1}{c|}{\multirow{-2}{*}{0.60}} & \multicolumn{1}{c|}{\multirow{-2}{*}{{\color[HTML]{FF0000} \textbf{0.80}}}} & \multicolumn{1}{c|}{\multirow{-2}{*}{0.65}} & \multicolumn{1}{c|}{\multirow{-2}{*}{0.61}} & \multirow{-2}{*}{0.28} \\ \hline
\end{tabular}}
\end{scriptsize}
 \vspace{-2 em}
\end{table}

\begin{figure}[pt]
	\centering
	\subfigure{		\includegraphics[width=0.46\linewidth]{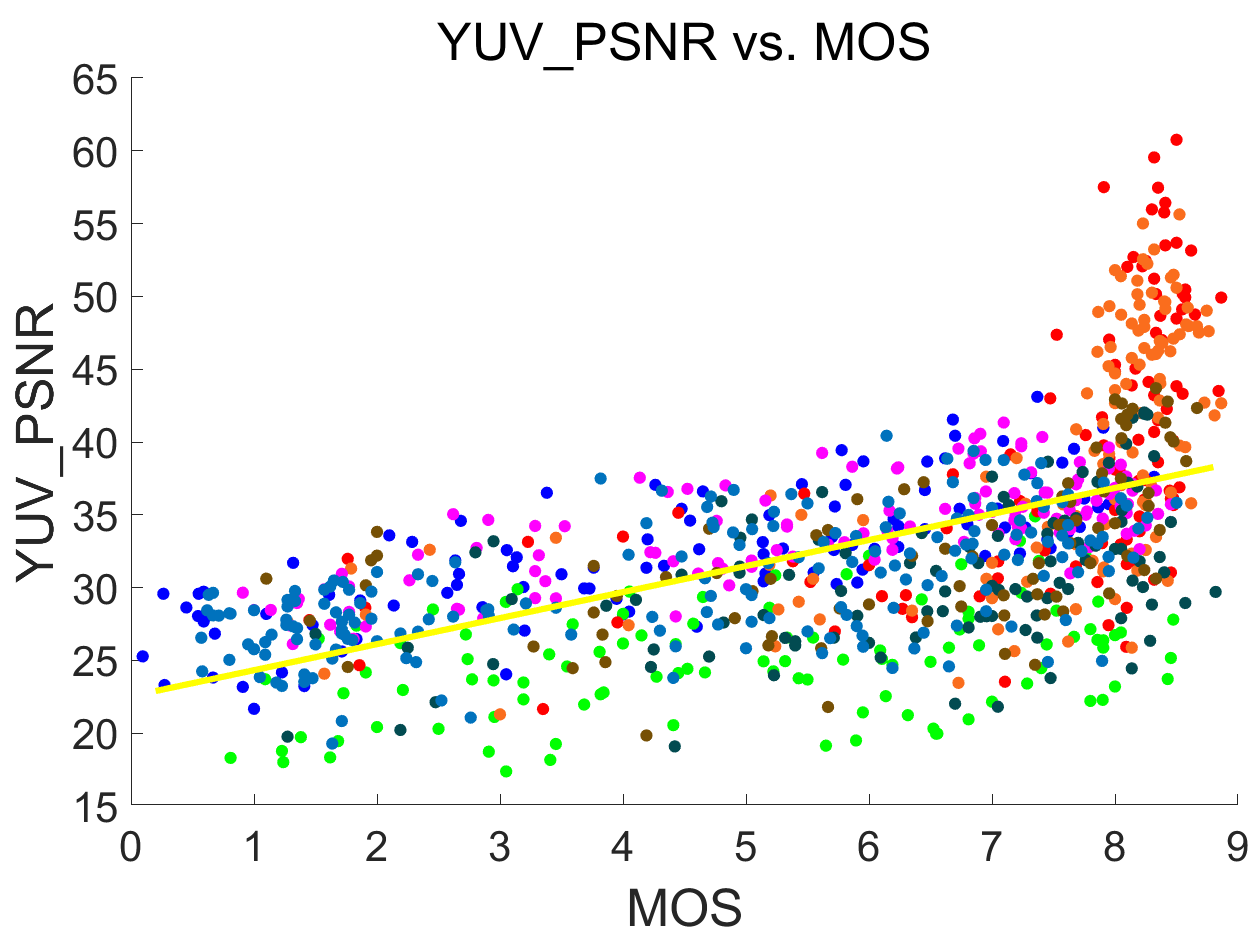}			}
	\subfigure{		\includegraphics[width=0.46\linewidth]{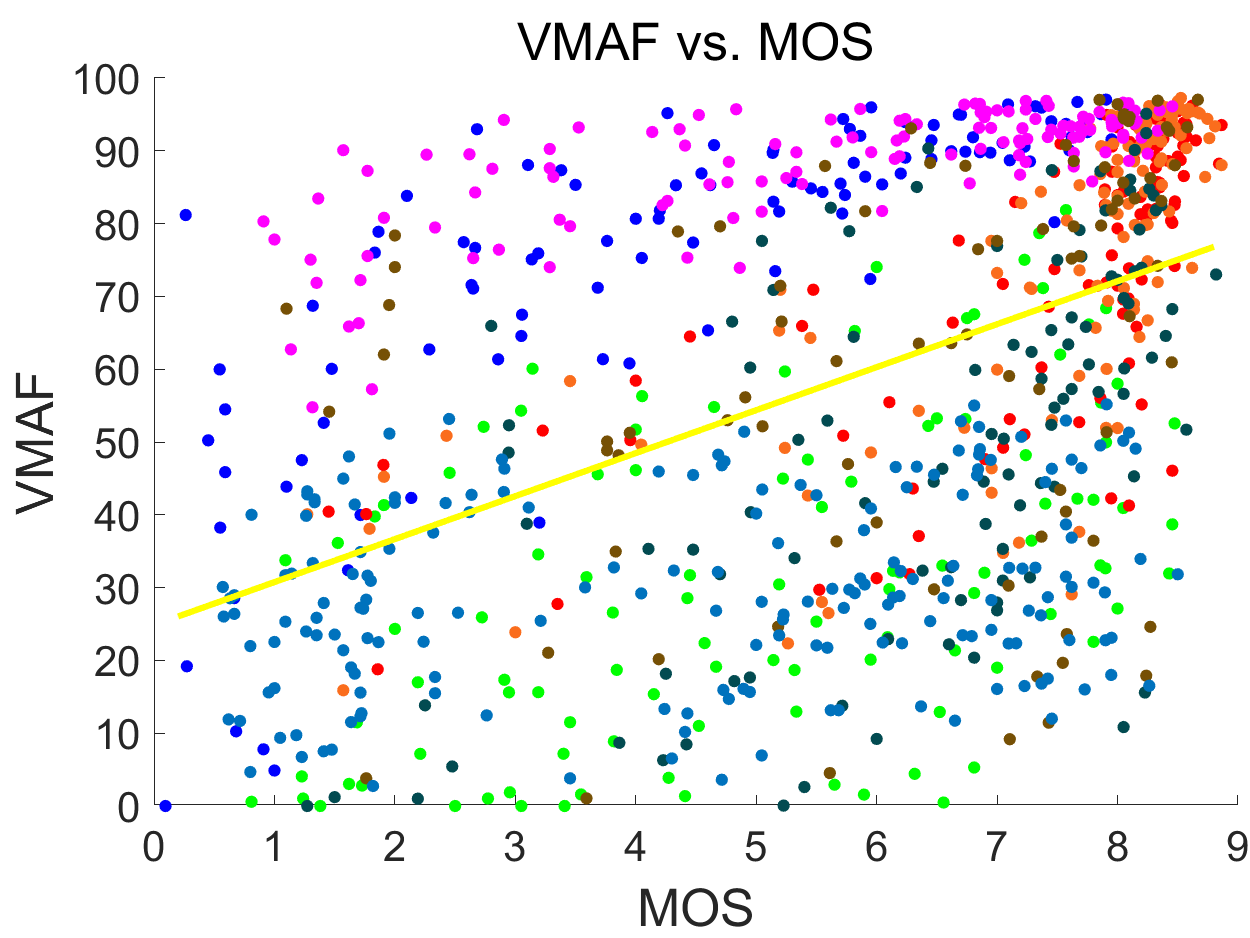}			}\\
         \subfigure{		\includegraphics[width=0.8\linewidth]{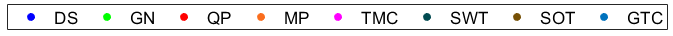}			}
	\caption{Scatter plot of objective metrics vs. MOS.}
	\label{fig:scatterplot}
  \vspace{-1.8em}
\end{figure}

The results of the metric on the entire database are shown in Table \ref{tb:metrics_performance} ``All'' columns. YUV\_PSNR reports the best performance, followed by RGB\_PSNR, PCQM\_PSNR, and VMAF. Fig. \ref{fig:scatterplot} shows the scatter plots of two metrics, in which the yellow lines represent the best-fitted cruves. We observe that the scatter plot of YUV\_PSNR is obviously better than VMAF, in which the scatter points are closer to the best-fit line. YUV\_PSNR tend to give low scores for GN samples. VMAF leans towards reporting high scores for QP and TMC.

The best overall correlations are below 0.6, which is far from the expectation that a robust metric should present a correlation at least above 0.80, indicating that SJTU-TMQA is a challenging database. Geo\_PSNR, D1, D2, and all model-based metrics show extremely low performance. The reason is that they only consider geometric features, while some samples in SJTU-TMQA are lossless with regard to geometry information, such as DS and TMC. 

\subsection{Analysis by type of distortion}
For an in-depth analysis,  the SRCC results for different types of distortion are illustrated in Table \ref{tb:metrics_performance} ``Distortion'' columns.  '-' means that the results of the metric for the samples applied with this kind of distortion are meaningless. VMAF presents good performance on DS distortion, in which it reports a correlation around 0.85. TPDM shows the best performance on GN and SWT with SRCC = 0.77 and 0.80. VMAF again exhibits the best performance on TMC, but the correlation is only 0.65. D1 and D2 showcase best results on QP and MQ with SRCC around 0.75 and 0.80, indicating that D1 and D2 might be good at predicting quantization distortion. PCQM\_PSNR reports a correlation around 0.70 on SOT, which is obviously better than most metrics. GTC is the most challenging type of distortion, in which no metric reports a correlation higher than 0.6. 

\subsection{Weakness of SOTA metrics}

Given that the highest correlation of the SOTA metric is only around 0.6, revealing that the SOTA metrics have weaknesses which are summarized as follows: for image and video-based metrics, one weakness is that projection might cause information loss \cite{yang2020predicting} and mask original mesh distortion. Furthermore, their performance is influenced by background information, which causes unstable score magnitudes for different types of contents \cite{tsmd}. For point-based metrics, the performance is closely related to the mesh sampling method. For the same mesh, different sampling methods and sampling resolutions can generate point clouds with obviously different perceptions, and consequently incur unstable metric performance \cite{yang2023tdmd}. For model-based metrics, most of them do not consider color attributes and cannot deal with geometry lossless distortion. Besides, they have strict requirements for tested meshes, such as the same connectivity or the same vertex density between reference and distorted meshes \cite{wg7}.

\section{Conclusion}\label{6}
In this paper, we create a large-scale textured mesh database called SJTU-TMQA which consists of 21 static textured meshes with diverse contents, rich distortion types, and accurate MOS.  The relationship between MOS and distortion is analyzed, and four types of SOTA objective metrics are evaluated based on SJTU-TMQA. The results demonstrate that human perception is influenced by content characteristics and distortion types,  and the best metric only achieves a correlation of around 0.60. This database can serve as a benchmark for objective metrics testing, providing opportunities for further metric research.

\bibliographystyle{IEEEbib}
\bibliography{mybib}
\end{document}